\documentclass[fleqn,10pt]{wlscirep}
\usepackage[utf8]{inputenc}
\usepackage{threeparttable}
\usepackage[T1]{fontenc}
\usepackage{graphicx}
\usepackage{longtable}
\usepackage{makecell}
\usepackage{subcaption}
\usepackage{makecell}
\title{Retrospective Comparative Analysis of Prostate Cancer In-Basket Messages: Responses from Closed-Domain LLM vs. Clinical Teams}

\author[1]{Yuexing Hao}
\author[1]{Jason M. Holmes}
\author[2]{Jared Hobson}
\author[2]{Alexandra Bennett}
\author[2]{Daniel K. Ebner}
\author[2]{David M. Routman}
\author[2]{Satomi Shiraishi}
\author[1]{Samir H. Patel}
\author[1]{Nathan Y. Yu}
\author[2]{Chris L. Hallemeier}
\author[2]{Brooke E. Ball}
\author[2]{Mark R. Waddle}
\author[1]{Wei Liu}
\affil [1]{Department of Radiation Oncology, Mayo Clinic Phoenix, AZ, 85054, USA}
\affil [2]{Department of Radiation Oncology, Mayo Clinic Rochester, MN, 55905, USA}

\begin{abstract}
In-basket message interactions play a crucial role in physician-patient communication, occurring during all phases (pre-, during, and post) of a patient's care journey. However, responding to these patients' inquiries has become a significant burden on healthcare workflows, consuming considerable time for clinical care teams. To address this, we introduce RadOnc-GPT, a specialized Large Language Model (LLM) powered by GPT-4 that has been designed with a focus on radiotherapeutic treatment of prostate cancer with advanced prompt engineering, and specifically designed to assist in generating responses. We integrated RadOnc-GPT with patient electronic health records (EHR) from both the hospital-wide EHR database and an internal, radiation-oncology-specific database. RadOnc-GPT was evaluated on 158 previously recorded in-basket message interactions. Quantitative natural language processing (NLP) analysis and two grading studies with clinicians and nurses were used to assess RadOnc-GPT's responses. Our findings indicate that RadOnc-GPT slightly outperformed the clinical care team in "Clarity" and "Empathy," while achieving comparable scores in "Completeness" and "Correctness." RadOnc-GPT is estimated to save 5.2 minutes per message for nurses and 2.4 minutes for clinicians, from reading the inquiry to sending the response. Employing RadOnc-GPT for in-basket message draft generation has the potential to alleviate the workload of clinical care teams and reduce healthcare costs by producing high-quality, timely responses.
\end{abstract}


\begin{document}

\flushbottom
\maketitle
%
%
\thispagestyle{empty}

\section{Introduction}

In-Basket is the online portal messaging system integrated within Epic Applications functioning similarly to email for communication between patients and their clinical care team. In-basket messaging system is often used to exchange messages regarding patient concerns, appointments, and follow-up care, particularly when real-time communication is not possible. During or following treatment, patients may not always receive immediate support from their care team. Patients with limited clinical literacy and understanding still need to communicate with healthcare professionals for various needs, including disease monitoring, medication, appointments, and billing or insurance issues. In this context, the In-basket serves as a vital tool to bridge the communication gap between patients and clinical professionals \cite{han2019using}. 

However, clinical care teams struggle to draft responses on time due to the increasing complexity of patients' supportive care needs \cite{sandford2016tracking}. Several studies have shown that an increased workload from responding to in-basket messages can negatively impact clinicians' burnout rates and overall well-being \cite{tai2019physicians, baxter2022association, sandford2016tracking, overhage2020physician}. Further, patient messaging volumes increased by more than 50\% after COVID-19, placing an undue burden on clinical teams \cite{nath2021trends, holmgren2023association, lieu2019primary}. Though these added avenues of communication are beneficial, generally responding to these messages is non-reimbursable as well  \cite{adler2020electronic, ayers2023comparing}.

Since In-basket messages often contain important real-world concerns from patients, the text-based in-basket message dataset is valuable for demonstrating patient-centered interactions. We propose using large language models (LLMs), which are connected with the electrical health record (EHR) system, to provide timely and layman-friendly responses to various categories of In-basket message inquiries \cite{achiam2023gpt, matulis2023relief, chen2024effect, gandhi2023can, baxter2024generative, small2024large}. LLMs have already shown strong technical capabilities in clinical context learning, summarization, response generation, decision-support, and Q\&A \cite{eriksen2023use, nori2023capabilities, hao2024advancing, holmes2023evaluating, garcia2024artificial, rezayi2022, wu2024, LIAO2024128576, HOLMES2024, liu2023tailoring, LIU2023100045, Dai2023ChatAugLC, XIAO2024102204}. Here, we aim to evaluate the performance of LLM and clinical care teams in three key areas: 1) capturing and interpreting all sources of data, 2) generating personalized and prompt responses, and 3) upholding high clinical standards in terms of completeness, correctness, clarity, and empathy.

Rather than applying LLMs across all types of disease sites, we focused on prostate cancer patients who received treatment at the Mayo Clinic’s radiation oncology department. Specializing in a specific field allows the LLM to generate more accurate and relevant responses. We developed RadOnc-GPT, an OpenAI GPT-4o-powered LLM, which is integrated with EHR \cite{liu2023radoncgptlargelanguagemodel}. Since many in-basket messages require external context for proper understanding and interpretation, RadOnc-GPT can generate more personalized responses with greater details in a zero-shot without training. This approach helps save time for the clinical care team by reducing the need to consult multiple sources of information to draft a response.

\section{Method}
This was a retrospective study approved by the Institutional Review Board of the Mayo Clinic. Our study included patients with prostate cancer who were managed at Mayo Clinic (Rochester, MN) in the calendar years 2022-2024. RadOnc-GPT is a Retrieval-Augmented Generation (RAG) system that connects with both the hospital wide electrical medical record database, Epic, developed by Epic Systems, and the radiation oncology specific database, Aria, developed by Varian Medical Systems. The data RadOnc-GPT may access includes clinical notes, radiology notes, pathology notes, urology notes, radiology reports, radiation treatment details, diagnosis details, patient details (demographics), in-basket messages, and more. RadOnc-GPT is able to retrieve data by way of specifying the patient ID and which dataset to retreive to the backend system. Once retrieved, the data is inserted into the conversation history.

Subject demographic information retrieved from the EHR system included sex, age, race, ethnicity, preferred language, and the attending physician's name. Information collected from Aria included demographic information (sex, age, race, ethnicity, preferred language, and the attending physician's name), prostate cancer treatment-specific information (course description, plan intent, treatment orientation, radiation type, radiation oncology machine type, number of fractions, dose prescription, dose delivered, radiation technique, and treatment duration), and diagnosis details (cancer stage, ICD (International Classification of Diseases) diagnosis code and code type, onset date). Information collected from Epic included clinical notes, ordered by date. For RadOnc-GPT, the information retrieval order starts with patient demographic details, followed by treatment details, diagnosis details, and lastly, clinical notes.

\begin{figure}[ht]
  \centering
  \includegraphics[width=16cm]{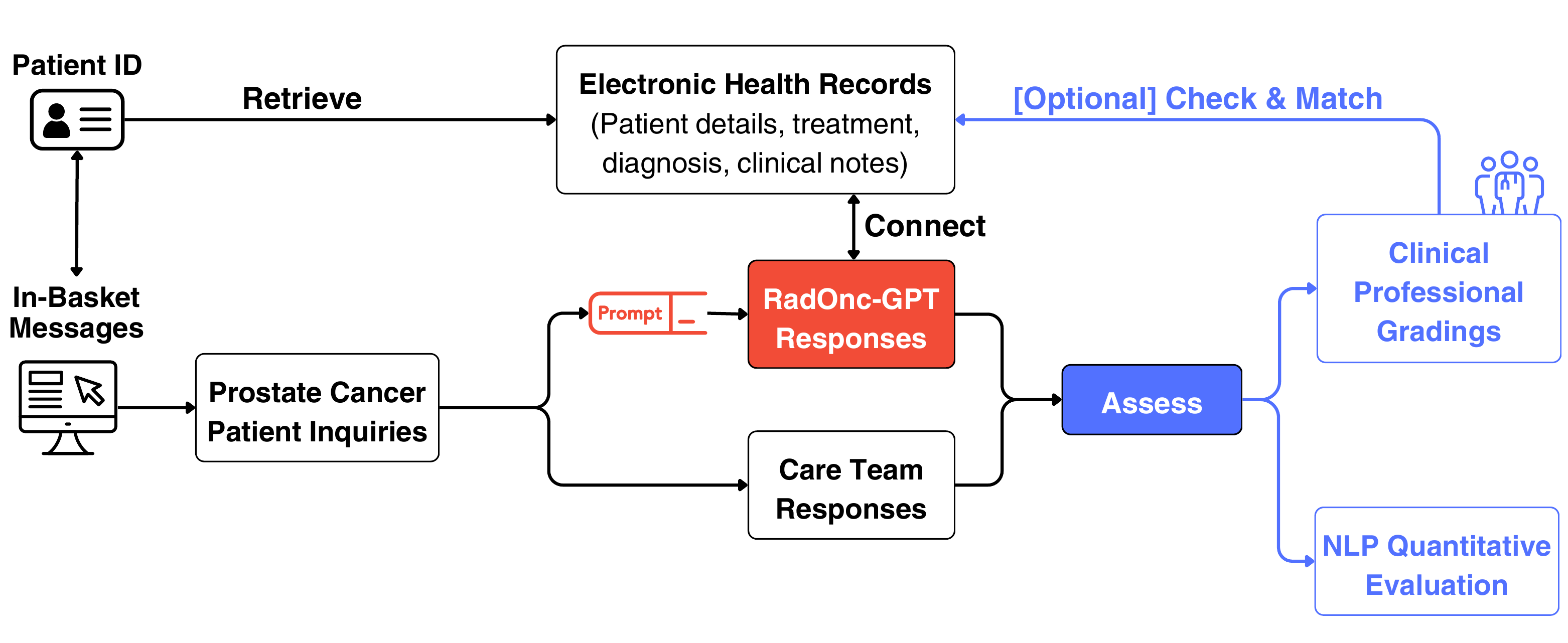}
  \caption{In-Basket Comparison Study Workflow Overview. From the in-basket messages dataset, we extracted prostate cancer patient inquiries and their corresponding care team responses. RadOnc-GPT, integrated with patients' EHR profiles, generates responses to these inquiries. A randomized dataset containing both RadOnc-GPT and clinical care team responses is then created for NLP-based quantitative evaluation and single-blinded grading by clinical professionals. Clinician and nurse graders can optionally review and match specific responses to patient EHRs using the patient ID.} 
  \label{Workflow}
\end{figure}

To ensure every patient inquiry was consistent and under the same GPT generation environment, we developed a GUI interface for RadOnc-GPT that was re-initialized for each test. This approach ensured that RadOnc-GPT did not generate biased responses from its memory of the previous patient's pair of inquiries and responses.  

Our study's evaluation was divided into two main components: natural language processing (NLP) quantitative assessments and clinical professional grading, as illustrated in Figure \ref{Workflow}. 

For NLP evaluation, we performed four types of measurements \cite{chang2024survey, iroju2015systematic}: natural language understanding, reasoning, context readability, and natural language generation. 

For the grading study, we focused on six dimensions of evaluation \cite{liu2024towards, abeysinghe2024challenges, wei2024evaluation}: 
\begin{enumerate}   
    \vspace{-2mm}
    \item completeness (ranging from 1-5, the higher the better),
    \vspace{-2mm}
    \item correctness (ranging from 1-5, the higher the better),
    \vspace{-2mm}
    \item clarity (ranging from 1-5, the higher the better), 
    \vspace{-2mm}
    \item empathy (ranging from 1-5, the higher the better), 
    \vspace{-2mm}
    \item estimated time to respond (in minutes),     \vspace{-2mm}
    \item extensive editing required (No use, major editing, minor editing, no editing needed),
    \vspace{-2mm}    
    \item (Optional) text comments section.
\end{enumerate}
   We enlisted four medical doctors from the Department of Radiation Oncology, all with significant in-basket response experience, with a mean Years of Experience (YoE) of 5. Of these, two medical residents (C1 and C2) independently graded all 158 messages. A third chief resident (C3) reviewed discrepancies when conflicts arose, and a board-certified radiation oncologist specializing in prostate cancer (C4) provided the final grading decision if disagreements persisted. Given that nurses typically initiate responses to in-basket messages, we also recruited four nurses from the same department to evaluate their capability (whether they can answer the questions or not) and estimate the time in minutes required to answer 158 patient inquiries (mean YoE = 5.25). The nurses provided anonymized estimates of the time in minutes spent responding to and redirecting these in-basket messages to other advanced practice providers (APPs) or clinicians. The graders details are displayed in Table \ref{tab:combined_clinician_nurse}.  Sample grading details are displayed in Appendix Figure \ref{Grading Study Details}.

\begin{table*}[ht] \centering \renewcommand{\arraystretch}{1.1}{ 
\centering \begin{tabular}{cccc|cccc} \toprule[2pt] \textbf{Clinician} & \textbf{Clinical Domain} & \textbf{Gender} & \textbf{YoE} & \textbf{Nurse} & \textbf{Cancer Domain}& \textbf{Gender} & \textbf{YoE} \\ \midrule[2pt] C1 & Radiation Oncology & Male & 3 yrs & N1 & Prostate \& Breast Cancer  & Female & 13 yrs \\ C2 & Radiation Oncology & Female & 3 yrs & N2 & Prostate Cancer  & Female & 4 yrs \\ C3 & Radiation Oncology & Male & 5 yrs & N3 & Prostate Cancer  & Female & 2 yrs \\ C4 & Radiation Oncology & Male & 9 yrs & N4 & Prostate Cancer  & Female & 2 yrs \\ \bottomrule[2pt] \end{tabular} } \vspace{0.2cm} \caption{Clinician and Nurse Grader Profiles.} \label{tab:combined_clinician_nurse} 
\end{table*}

\section{Results}
\subsection{In-Basket Message Dataset}
In-basket message interactions can often be disorganized. Without a standardized format for patient inquiries, under one subject, patients may send multiple messages for a single issue or combine several unrelated questions into one message. This makes it difficult to categorize the messages, as they frequently span multiple categories. Additionally, a single thread may include several conversation pairs, where a pair is defined as one or more patient inquiries followed by one or more care team responses in a time-sequenced manner. For inclusion in our evaluation dataset, each patient-clinician conversation pair must have consisted of one patient inquiry and one care team response.

\begin{figure}[h!]
  \centering
  \includegraphics[width=13cm]{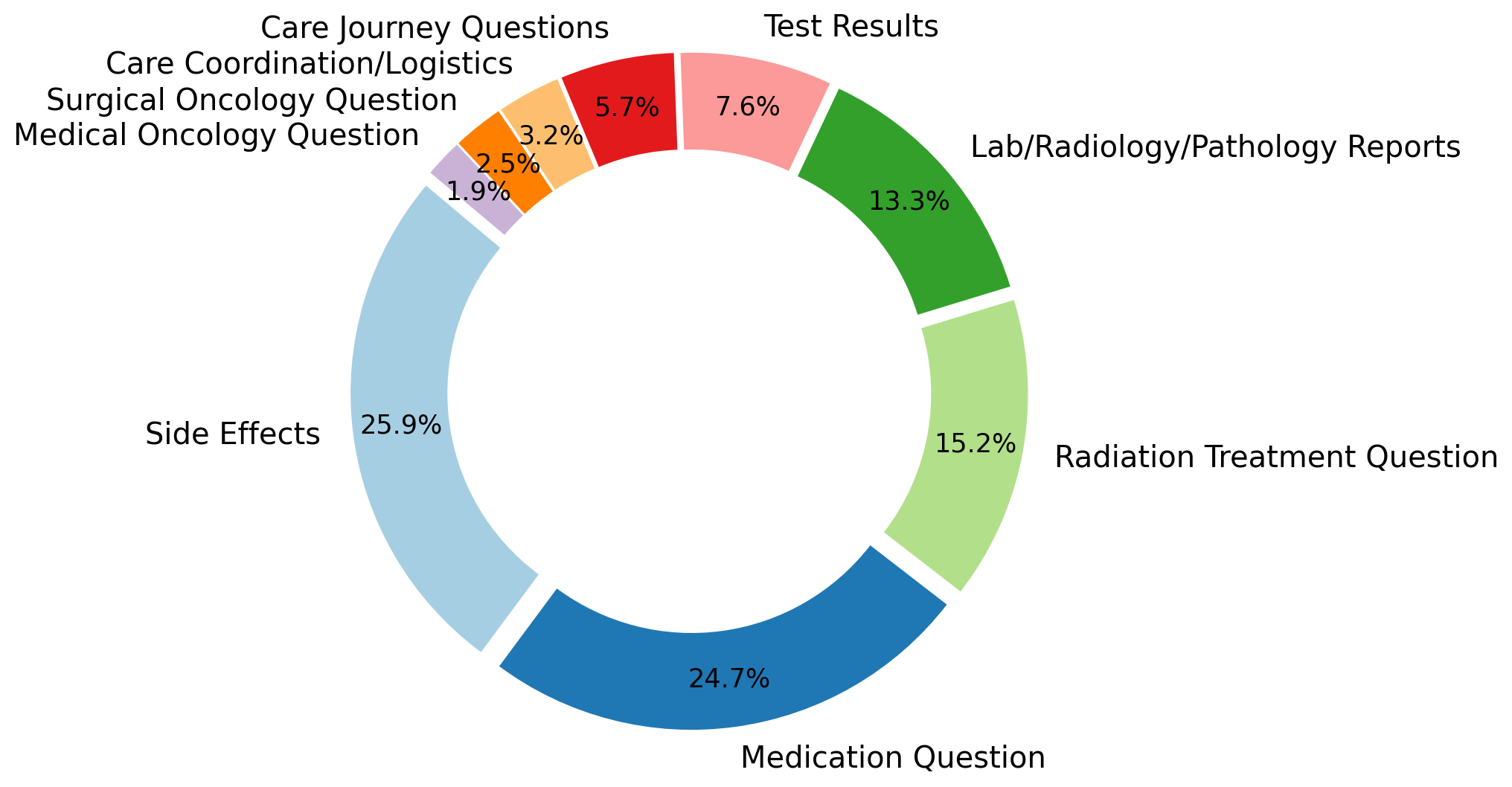}
  \caption{Nine Categories of In-Basket Message Patients' Inquiries}
  \label{Categories}
\end{figure}

We selected 90 non-metastatic prostate cancer patients from the Mayo Clinic Department of Radiation Oncology in-Basket message database. After filtering patient inquiries that are not relevant to medical advice seeking or receiving no or unrelated care team replies, we finally selected 158 patient inquiries, with each of them containing a clinical care team's reply. We only selected the message type under the Epic category of "Patient Medical Advice Request." We then pulled 158 patient inquiries' human care team's responses and utilized the patient inquiries to generate 158 RadOnc-GPT responses. We randomized the 316 responses and did not disclose the graders' response source. 

We manually summarized the 158 patient inquiries into 9 main categories: 'Test Results', 'Side Effects', 'Medication Questions', 'Radiation Treatment Questions', 'Medical Oncology Questions', 'Surgical Oncology Questions', 'Care Coordination/Logistics', 'Lab/Radiology/Pathology Reports', and 'Care Journey Questions' (Figure \ref{Categories}). The three most common patient inquiries are 'Side Effects', 'Medication Questions', and 'Radiation Treatment Questions'.

\subsection{NLP Analysis}
\subsubsection{Sentiment Analysis}
To understand the sentiment differences from human care team and the RadOnc-GPT, we conducted TextBlob and VADER analysis. In the TextBlob Sentiment Distribution (Figure \ref{NLP Analysis} (A)), RadOnc-GPT responses are observed to skew towards a more positive sentiment, with the majority of responses clustering around a sentiment score of 0.25. In contrast, human care team responses present a more evenly distributed sentiment profile, with a significant concentration around the neutral score of 0 (grey line in Figure \ref{NLP Analysis} (A)). RadOnc-GPT responses tend to be more positive, whereas Care Team responses consist of a broader spectrum of sentiments, including neutral and negative tones. The VADER Sentiment Distribution (Figure \ref{NLP Analysis} (B)) provides further insight into these differences. The box plot reveals that RadOnc-GPT responses exhibit a high median sentiment score, nearing 1.0, indicative of a predominantly positive sentiment. However, there are notable outliers reflecting occasional negative sentiment. Clinical care team responses, by comparison, display a wider range of sentiment scores, with a lower median, indicating a more varied and contextually nuanced sentiment expression. Our sentiment analysis collectively suggests that while RadOnc-GPT responses are generally more positive, care team responses offer a more balanced sentiment distribution, reflecting a greater sensitivity to the contextual nuances of the input data.

\subsubsection{Natural Language inference (NLI) analysis}

\begin{figure}[h!]
  \centering
  \includegraphics[width=18cm]{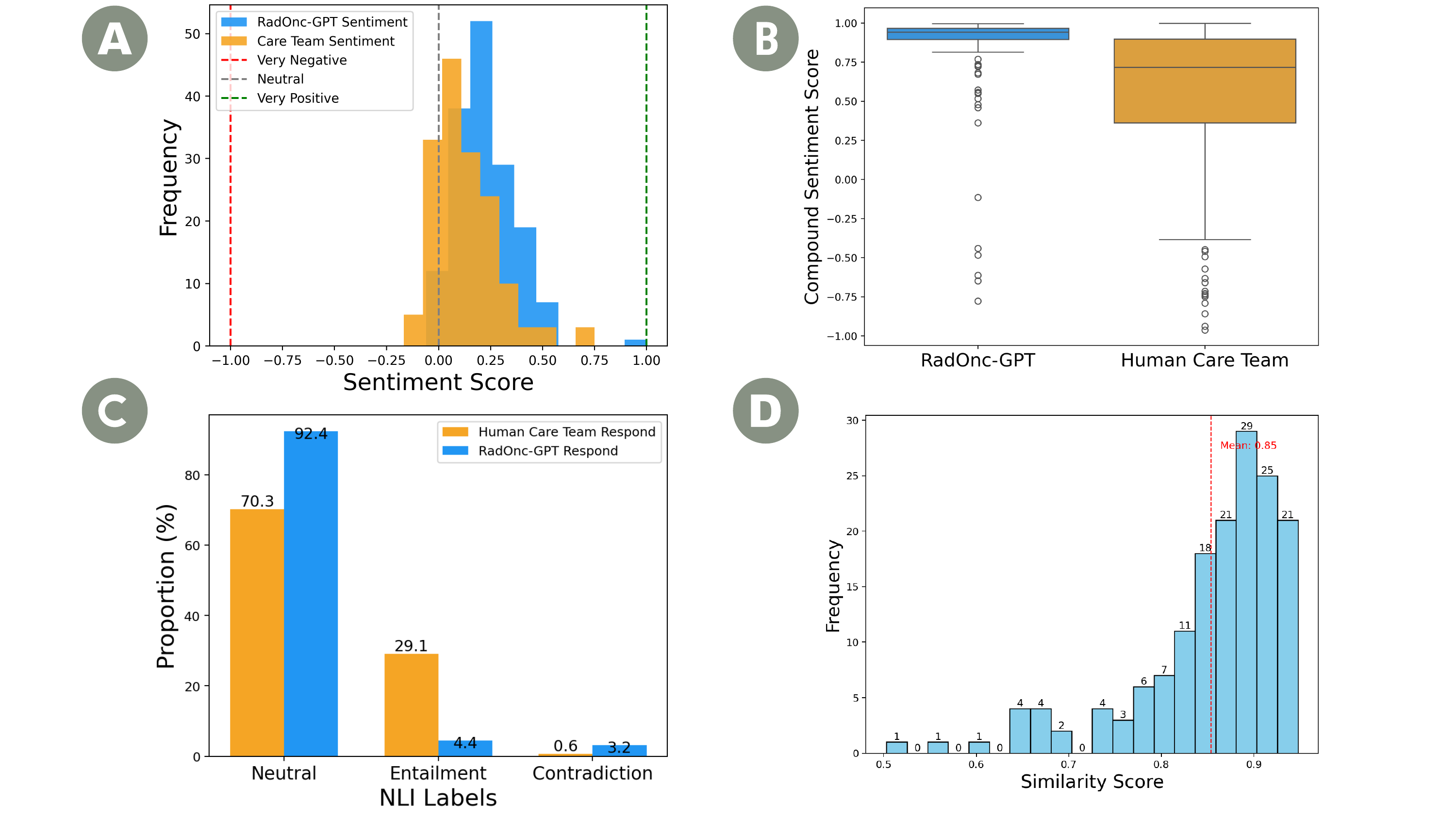}
  \caption{Sentiment Analysis. (A) TextBlob Sentiment Distribution; (B) VADER Sentiment Distribution; (C) NLI Distributions between GPT and Care Team Responses; (D) Semantic Similarity Scores}
  \label{NLP Analysis}
\end{figure}

To understand how human care team and RadOnc-GPT responses' inferences with the patients' inquiries, we conducted an NLI analysis \cite{maccartney2009natural}. RadOnc-GPT responses were predominantly Neutral, with 92.41\% of responses in this category, suggesting a tendency towards generalized statements. In contrast, clinician responses were more varied, with 70.25\% Neutral and 29.11\% Entailment, indicating greater relevance and specificity. Both response types showed low contradiction rates, though RadOnc-GPT responses had a slightly higher rate at 3.16\%, which may point to occasional inconsistencies. The NLI label distribution comparison is shown in Figure \ref{NLP Analysis} (C).

Comparing the semantic similarity \cite{miller1991contextual} between RadOnc-GPT and human care team responses provided additional context, showing a mean similarity score of 0.85 between RadOnc-GPT and human care team responses. This high score indicated a strong alignment in content, even though RadOnc-GPT responses are generally more neutral. The findings suggested that while RadOnc-GPT responses may lack the specificity found in human care team responses, they still captured the core semantic contents, reflecting contextually relevant information. Figure \ref{NLP Analysis} (D) shows the distribution of the semantic similarity scores.

\subsubsection{Readability Scores}

We compared the average readability scores across several indices, comparing patient inquiry, RadOnc-GPT, and clinical care team responses. The Flesch Reading Ease scores \cite{flesch1948new, jindal2017assessing}, where higher values indicate easier readability, showed that the human care team responses were the most accessible (66.2), followed by RadOnc-GPT (59.9). This suggested that human care team responses were slightly easier than RadOnc-GPT to read. For the Flesch-Kincaid Grade Level, Gunning Fog Index, SMOG Index, Automated Readability Index, and Coleman-Liau Index, lower scores indicated better readability. Across these metrics, human care team responses consistently scores lower than RadOnc-GPT, implying that human care team are written at a lower reading level and are easier to understand. RadOnc-GPT and Human Care Team responses, while similar across these indices, generally reflect higher complexity, particularly in the SMOG Index and Gunning Fog Index.

\begin{figure}[h!]
    \centering
    \includegraphics[width=18cm]{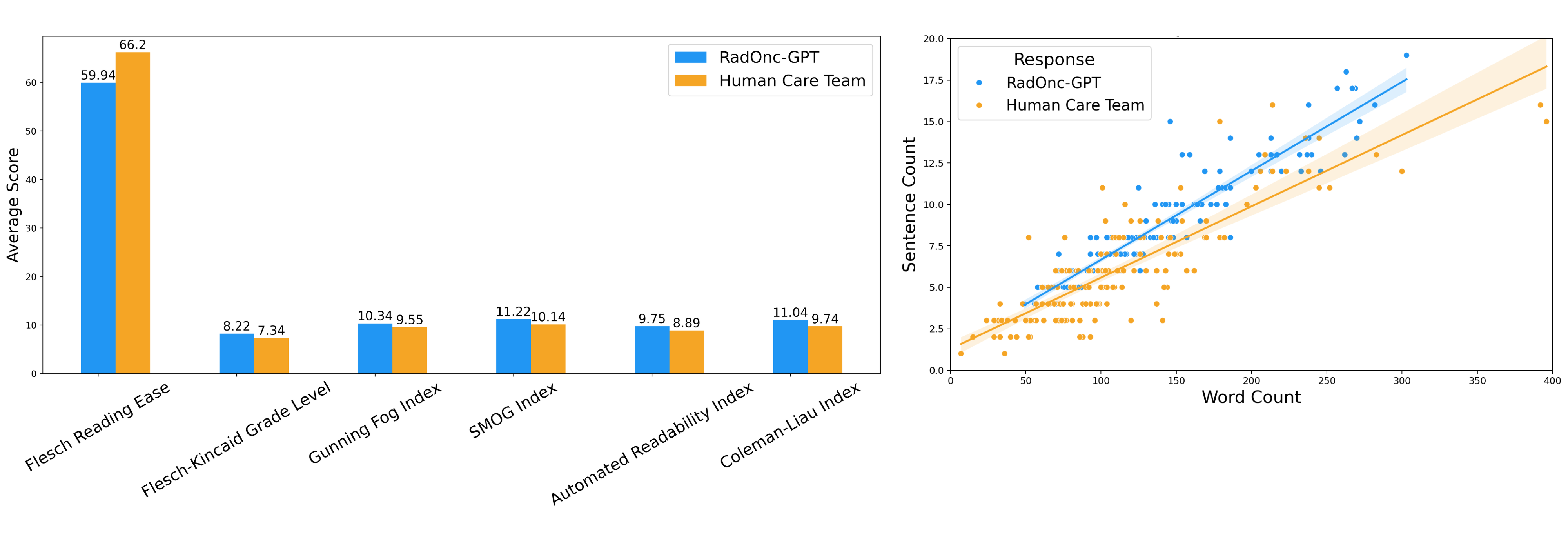}
    \caption{Readability Scores and Word and Sentence Counts Comparison.}
    \label{combined_Readability}
\end{figure}

The relationship between word counts and sentence counts in RadOnc-GPT, and human care team responses exhibited a positive correlation (Figure \ref{combined_Readability}). RadOnc-GPT responses tended to use more words per sentence than clinic care team responses. The steeper slope of the RadOnc-GPT regression line indicated that RadOnc-GPT responses became more verbose as the sentence count increases. Human care team responses were more clustered at lower word and sentence counts, reflecting a more concise communication style. Figure \ref{combined_Readability} clearly distinguished GPT's verbosity from the clinic care team's brevity.

On average, RadOnc-GPT responses were more detailed, with about 135 words and 9 sentences per response. Human care team responses, while similar in length to RadOnc-GPT responses, average around 132 words and 7 sentences per response, indicating that care team responses were slightly more concise in terms of sentence structure.



\subsection{Clinician Grader Study}

In the single-blinded grader study, two clinician graders first graded all 316 responses (158 human care team responses, 158 RadOnc-GPT responses). The average of two graders' results showed that GPT consistently performs better in "\textit{Empathy}" and "\textit{Clarity}', while human responses show higher averages in "\textit{Completeness}" and "\textit{Correctness}". The grading rubrics are dispalyed in Appendix \ref{rubric}.

\begin{figure}[h!]
    \centering
    \includegraphics[width=12cm]{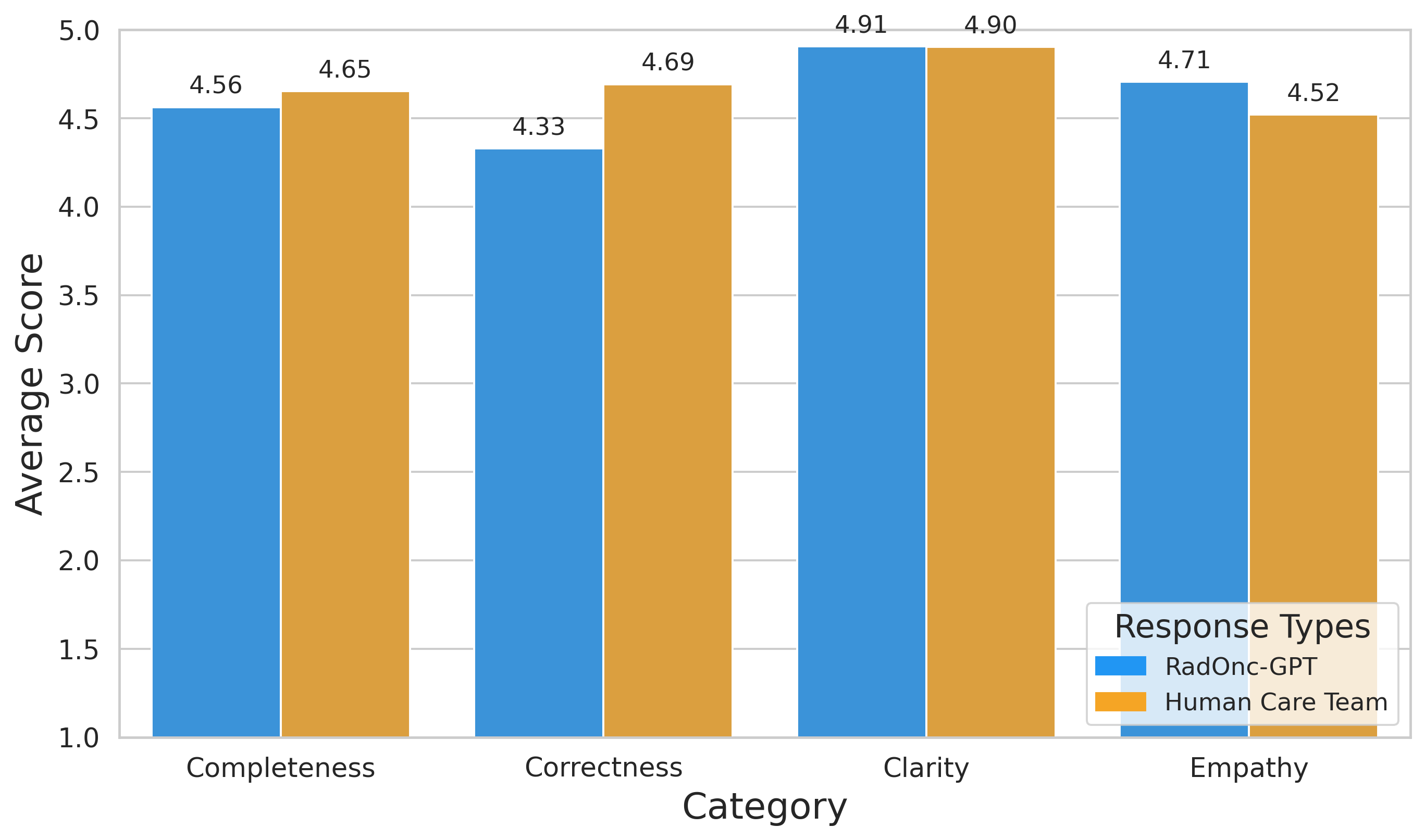}
    \caption{Average Score Across all Four Categories}
    \label{fig:Time}
\end{figure}

\begin{figure}[h!]
  \centering
  \includegraphics[width=11cm]{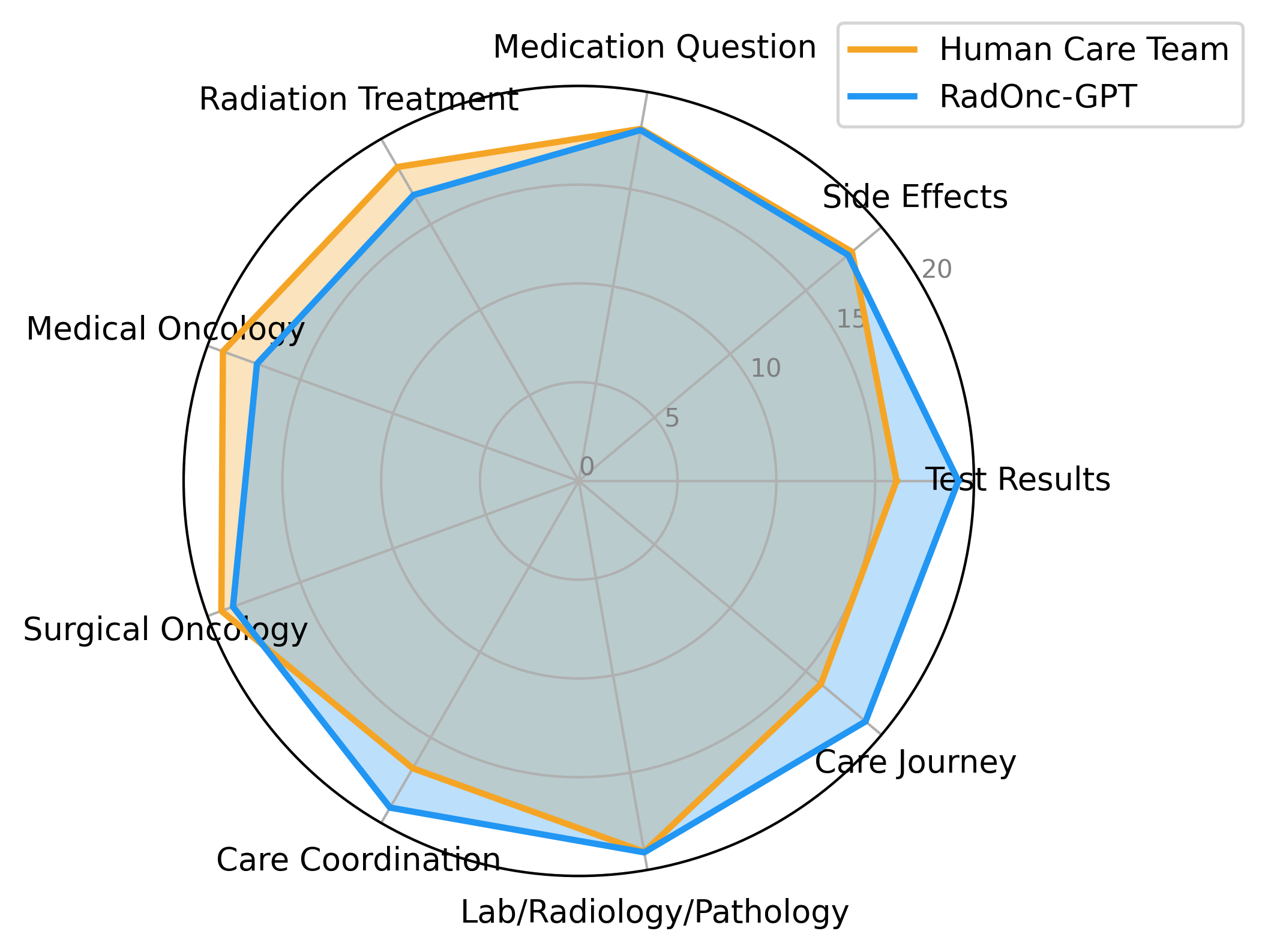}
  \caption{Radar Chart Comparing Clinical Care Team and RadOnc-GPT Performance Across Nine Categories. The four grading dimensions are combined, with a maximum possible score of 20.}
  \label{Category_Scores}
\end{figure}

The mean clinician scores on "\textit{Completeness}," "\textit{Correctness}," "\textit{Clarity}," and "\textit{Empathy}" for "Human" responses vary between 16.00 and 19.25 across categories, with the highest in Surgical Oncology Question (19.25). In contrast, RadOnc-GPT scores range from 16.72 to 19.21, with the highest in Test Results (19.21). RadOnc-GPT outperforms clinical care team in \textbf{Test Results}, \textbf{Care Coordination/Logistics}, and \textbf{Care Journey Questions}, as displayed in Figure \ref{Category_Scores}.

\subsubsection{Time Comparison}

\begin{figure}[h!]
  \centering
  \includegraphics[width=18cm]{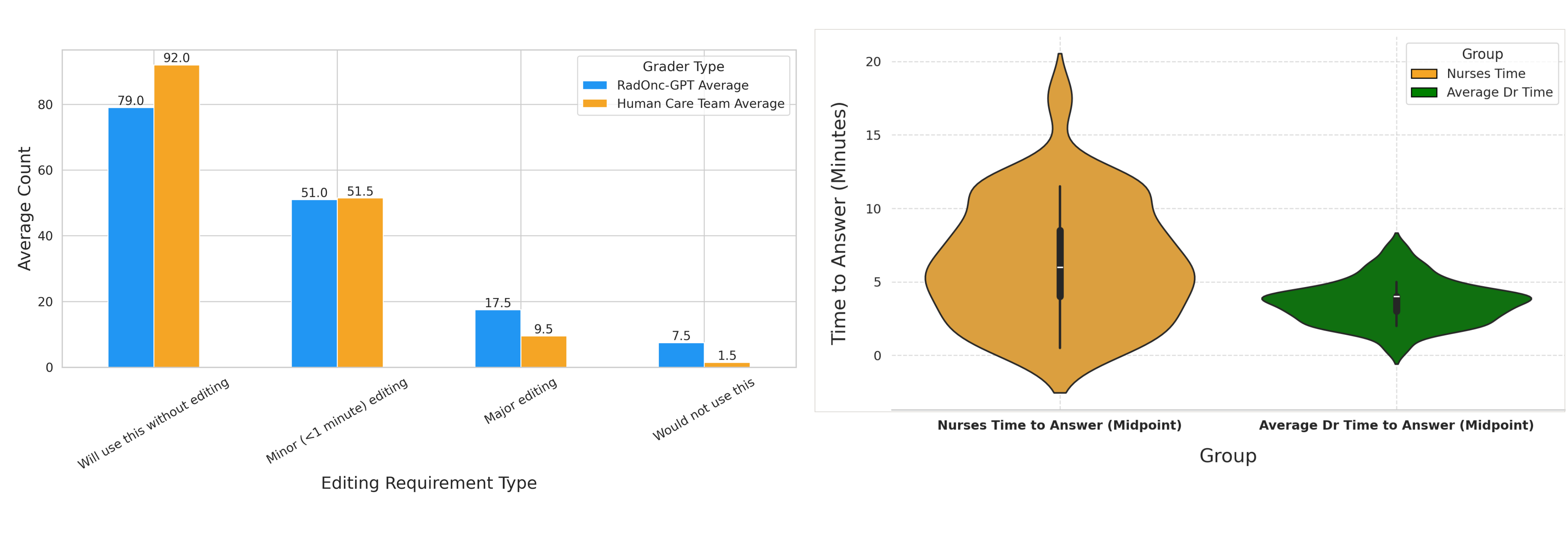}
  \caption{[Left] Comparisons on Two Graders' Average Human Care Team and RadOncGPT Editing Time; [Right] Comparative Analysis of Clinicians and Nurses Average Time in Responding In-Basket Messages.}
  \label{Time_Comparison}
\end{figure}


The nurse graders study focused solely on two criteria: "\textit{Can you answer this patient inquiry?}" and "\textit{Estimated time to answer this patient inquiry}." We compared the clinician graders' estimated times to those of the nurses. On average, clinicians took 3.60 minutes (SD 1.44) to respond to an in-basket message, compared to the nurses' 6.39 minutes (SD 4.05). While both clinician graders were able to answer all 158 messages, nurses indicated "No" for 90 inquiries, requiring referral to clinicians, and "Yes" for 68 inquiries. For the inquiries marked "Yes," the average response time was 5.54 minutes, and for those marked "No," the average time was 8.83 minutes. Even though nurses may struggle with some inquiries, they still need to conduct proper research and gather relevant patient information to determine whether the in-basket message should be escalated to clinicians.

\section{Discussion}

RadOnc-GPT was well able to provide medical advice to individualized patient In-basket messages on this retrospective comparison study to both trained radiation oncologists as well as radiation-oncology-specific nurses.  Although RadOnc-GPT responses are human-like and generally similar to responses generated by the original human care teams in many aspects, caution is still needed before deploying its messages without human oversight in real-world healthcare settings. 

\begin{figure}[h!]
  \centering
  \includegraphics[width=13cm]{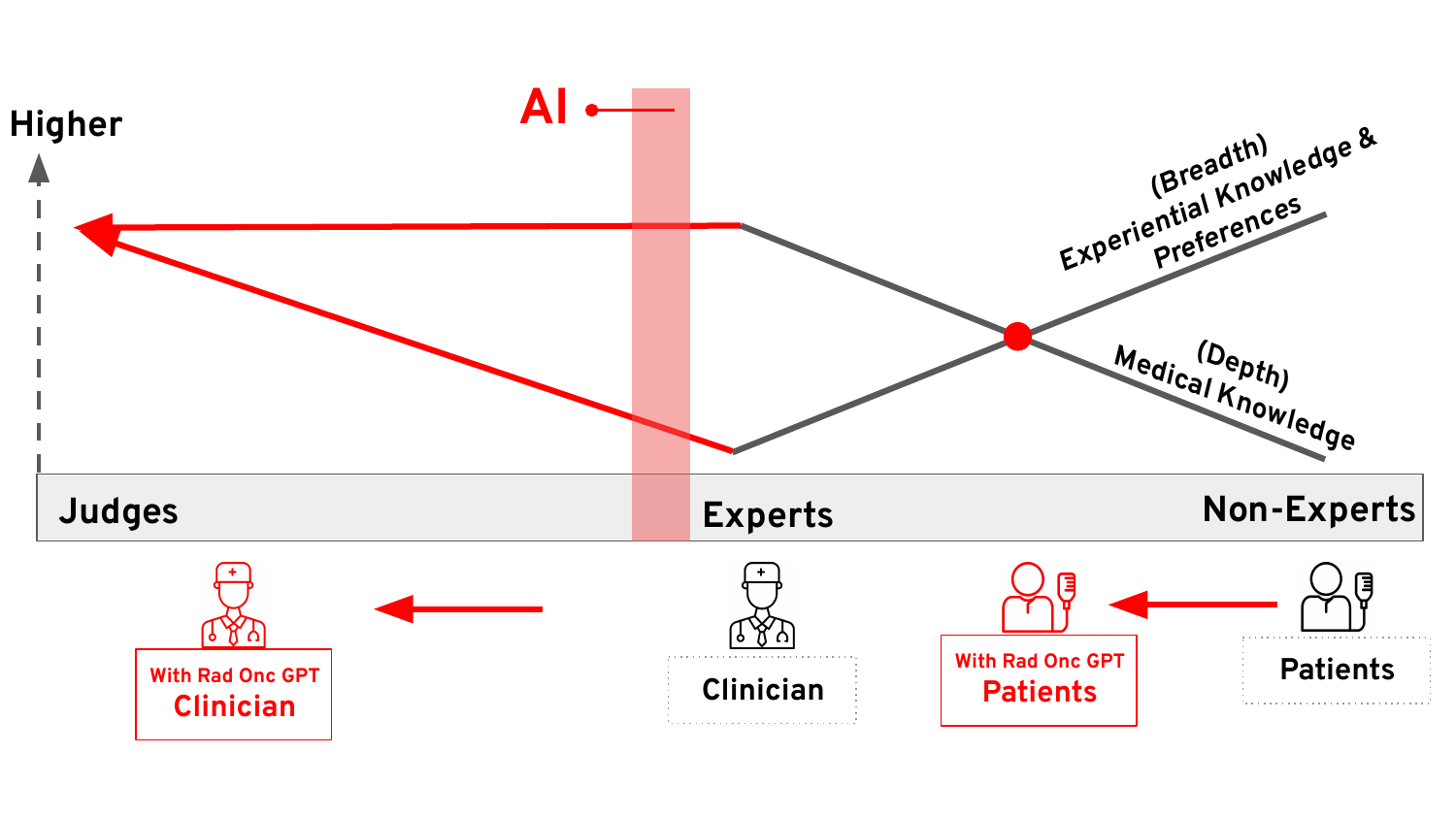}
  \caption{Patient-Clinician Hierarchy Structure Shift with RadOncGPT in In-Basket Message Generation. With RadOnc-GPT's assistance in in-basket message generation, human care team can gradually switch its roles in initiating the response drafts to judging the RadOnc-GPT generated drafts.}
  \label{Structure}
\end{figure}

 Since human care team may still need to confirm the evidence in responses by pulling out the imaging or lab/exam results to avoid hallucination, RadOnc-GPT may be able to accelerate the response turn-out rate and alleviate the human care team's response pressure. 

Our study observed that the human care team responses typically addressed the immediate action items to instruct patients what to do next. The care team seldom provides sufficient patient education, clinical concepts clarifications, or informed explanations. As RadOnc-GPT responses provided more information that clinical care team's responses might not include, RadOnc-GPT pushed non-expert patients to gain more expertise. While RadOnc-GPT prepared a draft in-basket message response, clinicians went from \textit{Experts} to \textit{Judges}. The shift of both patients' and clinicians' roles and expertise in healthcare was illustrated in Figure \ref{Structure}.

\subsection{Prompt Engineering}
We considered prompts to be one of the key factors determining the quality of RadOnc-GPT responses. For the final prompts, we provided instructions in 1) steps of retrieving information to ensure responsibility; 2) acting as the attending physician and provider; 3) step-by-step reasoning from patient health profiles to address patient's inquiries; 4) handling the medications (prioritizing over-the-counter medications); 5) determining the clarity of patient's inquiry and asking for more information if needed; 6) patient's health literacy; 7) providing the original patient's inquiry. The full prompts were presented in the appendix \ref{prompt}.

Additionally, there is a lack of standardized scales or metrics to evaluate the GPT-generated messages. A few studies have included clear evaluation methods and scoring rubrics for grading. However, the studies in the medical domain are quite specific, and researchers found it challenging to generalize the grading across all types of medical domains or diseases. Also, in the reader study, which included human evaluators, the subjective grading could potentially introduce bias from years of experience, practicing domain, clinical roles, and clinics. 

\subsection{Economic Potential Impacts}

\begin{figure}[h!]
  \centering
  \includegraphics[width=13cm]{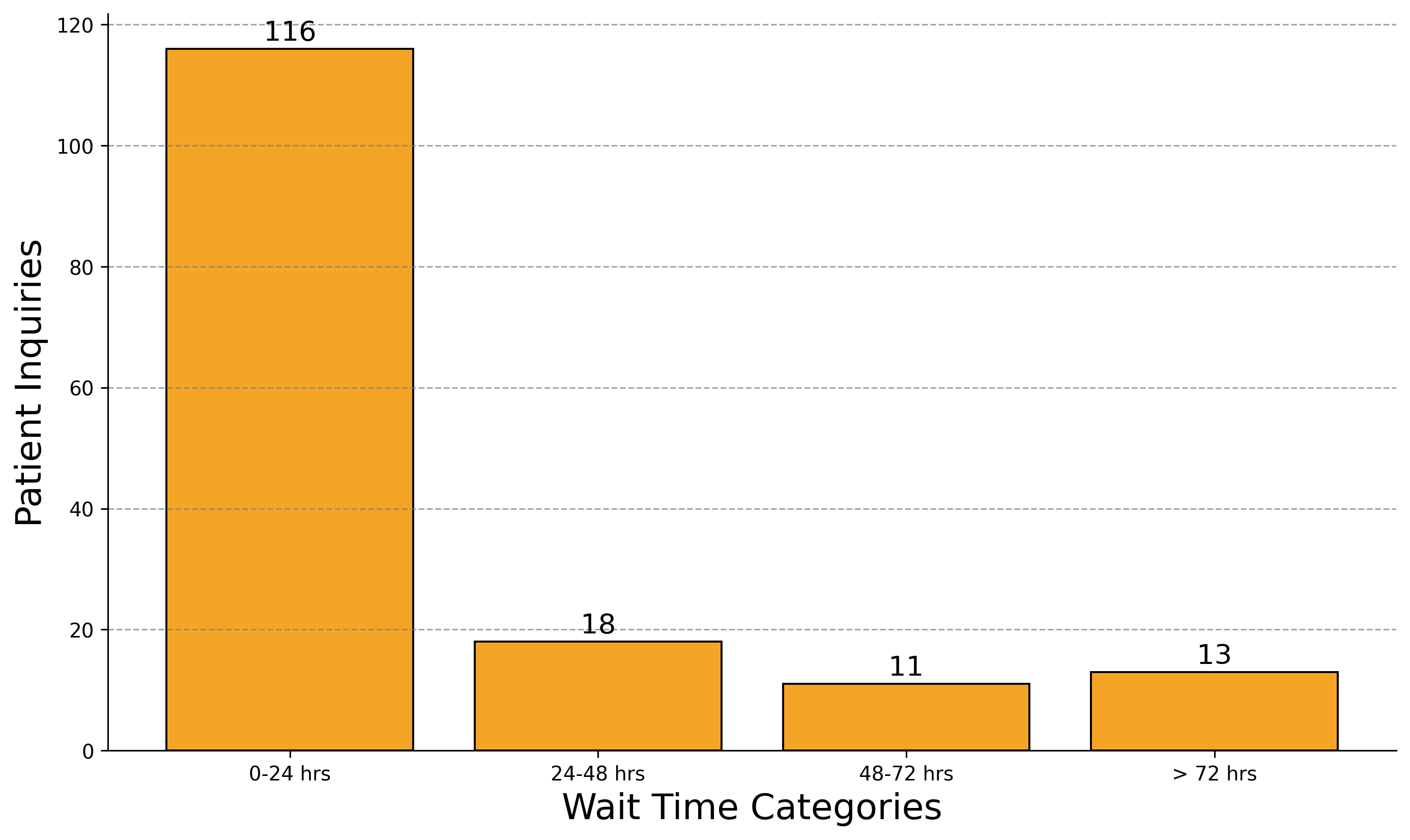}
  \caption{Wait Time for Human Care Team's Response}
  \label{Wait Time}
\end{figure}

The average of this 158 patient inquiry messages wait time for clinical care team response is 22.42 hrs (sd = 32.83, median = 11.73 hrs), as shown in Figure \ref{Wait Time}.

The purpose of using RadOnc-GPT to generate in-basket message response was not to replace the human care team's role in managing the prostate cancer patients' inquiries. Instead, RadOnc-GPT was intended to streamline the response process and save time for the care team. Typically, responses to in-basket messages were handled sequentially, starting with nurses, then progressing to nurse practitioners or APPs, and finally to clinicians.

Based on our estimation, using RadOnc-GPT to assist in in-basket messages generation, average words in patient inquiries were 88.89 (SD: 64.93), estimated reading time (for an average English reader 175 words per min) would be 0.51 min for each message (SD: 0.37 min). GPT response average words were 119.55 (SD: 49.72), with an estimated reading time of 0.68 min (SD: 0.28 min). The clinical professionals review time would be 1.19 min for each message. Based on the clinicians and nurses estimation, RadOnc-GPT could save approximately 5.2 minutes per message for nurses and 2.41 minutes for clinicians, from reading the patient inquiry mesage to drafting and sending the response. With Mayo Clinic receiving around 5,000 in-basket messages daily \cite{cognetta_rieke_2024} and assuming that one-fifth of these are requests for medical advice (which is 1000 messages), the potential time savings for nurses alone would amount to 5200 minutes (or 86.67 hours) per day. Based on the NIH salary table\footnotemark, this equates to an annual savings of at least \$2.28 million in nurse time (\$72 per hour).
\footnote{https://hr.nih.gov/benefits/pay/pay-guide}

\section{Limitation}
The retrospective study feature limited our study since we can't ask the patients to add more information or reply to the RadOnc-GPT generated responses. We only compared a pair of interactions under one subject, which consists of a patient inquiry and a response message from either RadOnc-GPT or the clinical care team. It might deviate from the real-world interaction since sometimes either the clinical care team or patients send out multiple messages under one subject to explain their health concerns. Additionally, RadOnc-GPT processes only text and is currently unable to handle images or files. While no images or files were involved in the in-basket message grading, interpreting such information typically takes longer than reading text messages. 

Although RadOnc-GPT generated responses can be comparable to clinical care team responses, this was a retrospective study, and the human care team's responses were affected by multiple factors (i.e., different care team roles' response, the busy time, clinical department), and likely the responses were not the best from the clinical care team. 

Another limitation is that we only use GPT-4o as the backend LLM for RadOnc-GPT to generate responses. We didn't compare our backend engine GPT-4o with other LLMs such as LLama 3, Gemini, GPT-4 or GPT-3.5. The performance based on GPT-4o may not generalize to other LLMs. 

\section{Conclusion}
In this single-blinded comparison study, we evaluated 158 in-basket message interactions between RadOnc-GPT and clinical care teams. The results demonstrated RadOnc-GPT's ability to answer patient inquiries, though we observed limitations in its capacity to capture the nuanced information that clinical professionals provide. Utilizing RadOnc-GPT as a foundational tool for generating in-basket message responses allows clinical professionals to serve more as reviewers than primary authors. This approach not only saves time and improves workflow efficiency but also enables clinicians to be more comprehensive in their responses and to focus more on the direct patient interaction care. Future studies should further explore the limitations of LLMs in assisting with in-basket message generation.

\section{Data Availability}
The authors declare that the data supporting the findings of this study are available upon request. The dataset is not public because it contains patient health information (PHI). However, sample data consisting of 18 pairs of patient inquiries and responses, with PHI removed, is available on GitHub: \href{https://github.com/YuexingHao/In-Basket-Message-Evaluation/blob/main/In-Basket-QA-Dataset.xlsx}{https://github.com/YuexingHao/In-Basket-Message-Evaluation/blob/main/In-Basket-QA-Dataset.xlsx}.

\section{Code Availability}
The code is available on GitHub: 
\href{https://github.com/YuexingHao/In-Basket-Message-Evaluation}{https://github.com/YuexingHao/In-Basket-Message-Evaluation}

\section{Acknowledgments}
We thank the nurse practitioners Derek S. Remme, D.N.P., and Jonathan Moonen, D.N.P., as well as the nurses who contributed to this study: Brittainy Johnson, R.N., Shyanne Dobbs, R.N., Brooke Kelly, R.N., and Bailey Kirchner, R.N. This research was supported by the National Cancer Institute (NCI) R01CA280134, the Eric \& Wendy Schmidt Fund for AI Research \& Innovation, the Fred C. and Katherine B. Anderson Foundation, and the Kemper Marley Foundation. The authors also acknowledged support from Paul Calabresi Program in Clinical/Translational Research at the Mayo Clinic Comprehensive Cancer Center K12CA090628.

\section{Author Contributions}
Y.H., J.M.H., M.R.W., and W.L. conceptualized the study. Y.H. and J.M.H. were responsible for data collection, data preprocessing, model development, and validation. J.H., A.B., D.K.E., S.S., S.H.P., B.E.B., C.L.H., and M.R.W. offered expertise in clinical grading studies and interpreted the results. Y.H. J.M.H., D.M.R., N.Y.Y., B.E.B., D.K.E., M.R.W., and W.L. interpreted the experimental results and provided feedback on the study. All authors contributed to writing the manuscript and reviewed and approved the final version. The study was co-supervised by M.R.W. and W.L.

\section{Competing Interests}
The authors declare no competing interests.

\bibliography{main}

\begin{thebibliography}{10}
\urlstyle{rm}
\expandafter\ifx\csname url\endcsname\relax
  \def\url#1{\texttt{#1}}\fi
\expandafter\ifx\csname urlprefix\endcsname\relax\def\urlprefix{URL }\fi
\expandafter\ifx\csname doiprefix\endcsname\relax\def\doiprefix{DOI: }\fi
\providecommand{\bibinfo}[2]{#2}
\providecommand{\eprint}[2][]{\url{#2}}

\bibitem{han2019using}
\bibinfo{author}{Han, H.-R.} \emph{et~al.}
\newblock \bibinfo{journal}{\bibinfo{title}{Using patient portals to improve patient outcomes: systematic review}}.
\newblock {\emph{\JournalTitle{JMIR human factors}}} \textbf{\bibinfo{volume}{6}}, \bibinfo{pages}{e15038} (\bibinfo{year}{2019}).

\bibitem{sandford2016tracking}
\bibinfo{author}{Sandford, L.~M.} \emph{et~al.}
\newblock \bibinfo{journal}{\bibinfo{title}{Tracking health care team response to electronic health record asynchronous alerts: Role of in-basket message burden}}.
\newblock {\emph{\JournalTitle{Journal of Patient-Centered Research and Reviews}}} \textbf{\bibinfo{volume}{3}}, \bibinfo{pages}{201--202} (\bibinfo{year}{2016}).

\bibitem{tai2019physicians}
\bibinfo{author}{Tai-Seale, M.} \emph{et~al.}
\newblock \bibinfo{journal}{\bibinfo{title}{Physicians’ well-being linked to in-basket messages generated by algorithms in electronic health records}}.
\newblock {\emph{\JournalTitle{Health Affairs}}} \textbf{\bibinfo{volume}{38}}, \bibinfo{pages}{1073--1078} (\bibinfo{year}{2019}).

\bibitem{baxter2022association}
\bibinfo{author}{Baxter, S.~L.} \emph{et~al.}
\newblock \bibinfo{journal}{\bibinfo{title}{Association of electronic health record inbasket message characteristics with physician burnout}}.
\newblock {\emph{\JournalTitle{JAMA Network Open}}} \textbf{\bibinfo{volume}{5}}, \bibinfo{pages}{e2244363--e2244363} (\bibinfo{year}{2022}).

\bibitem{overhage2020physician}
\bibinfo{author}{Overhage, J.~M.} \& \bibinfo{author}{McCallie~Jr, D.}
\newblock \bibinfo{journal}{\bibinfo{title}{Physician time spent using the electronic health record during outpatient encounters: a descriptive study}}.
\newblock {\emph{\JournalTitle{Annals of internal medicine}}} \textbf{\bibinfo{volume}{172}}, \bibinfo{pages}{169--174} (\bibinfo{year}{2020}).

\bibitem{nath2021trends}
\bibinfo{author}{Nath, B.} \emph{et~al.}
\newblock \bibinfo{journal}{\bibinfo{title}{Trends in electronic health record inbox messaging during the covid-19 pandemic in an ambulatory practice network in new england}}.
\newblock {\emph{\JournalTitle{JAMA network open}}} \textbf{\bibinfo{volume}{4}}, \bibinfo{pages}{e2131490--e2131490} (\bibinfo{year}{2021}).

\bibitem{holmgren2023association}
\bibinfo{author}{Holmgren, A.~J.}, \bibinfo{author}{Byron, M.~E.}, \bibinfo{author}{Grouse, C.~K.} \& \bibinfo{author}{Adler-Milstein, J.}
\newblock \bibinfo{journal}{\bibinfo{title}{Association between billing patient portal messages as e-visits and patient messaging volume}}.
\newblock {\emph{\JournalTitle{Jama}}} \textbf{\bibinfo{volume}{329}}, \bibinfo{pages}{339--342} (\bibinfo{year}{2023}).

\bibitem{lieu2019primary}
\bibinfo{author}{Lieu, T.~A.} \emph{et~al.}
\newblock \bibinfo{journal}{\bibinfo{title}{Primary care physicians’ experiences with and strategies for managing electronic messages}}.
\newblock {\emph{\JournalTitle{JAMA network open}}} \textbf{\bibinfo{volume}{2}}, \bibinfo{pages}{e1918287--e1918287} (\bibinfo{year}{2019}).

\bibitem{adler2020electronic}
\bibinfo{author}{Adler-Milstein, J.}, \bibinfo{author}{Zhao, W.}, \bibinfo{author}{Willard-Grace, R.}, \bibinfo{author}{Knox, M.} \& \bibinfo{author}{Grumbach, K.}
\newblock \bibinfo{journal}{\bibinfo{title}{Electronic health records and burnout: time spent on the electronic health record after hours and message volume associated with exhaustion but not with cynicism among primary care clinicians}}.
\newblock {\emph{\JournalTitle{Journal of the American Medical Informatics Association}}} \textbf{\bibinfo{volume}{27}}, \bibinfo{pages}{531--538} (\bibinfo{year}{2020}).

\bibitem{ayers2023comparing}
\bibinfo{author}{Ayers, J.~W.} \emph{et~al.}
\newblock \bibinfo{journal}{\bibinfo{title}{Comparing physician and artificial intelligence chatbot responses to patient questions posted to a public social media forum}}.
\newblock {\emph{\JournalTitle{JAMA internal medicine}}} \textbf{\bibinfo{volume}{183}}, \bibinfo{pages}{589--596} (\bibinfo{year}{2023}).

\bibitem{achiam2023gpt}
\bibinfo{author}{Achiam, J.} \emph{et~al.}
\newblock \bibinfo{journal}{\bibinfo{title}{Gpt-4 technical report}}.
\newblock {\emph{\JournalTitle{arXiv preprint arXiv:2303.08774}}}  (\bibinfo{year}{2023}).

\bibitem{matulis2023relief}
\bibinfo{author}{Matulis, J.} \& \bibinfo{author}{McCoy, R.}
\newblock \bibinfo{journal}{\bibinfo{title}{Relief in sight? chatbots, in-baskets, and the overwhelmed primary care clinician}}.
\newblock {\emph{\JournalTitle{Journal of general internal medicine}}} \textbf{\bibinfo{volume}{38}}, \bibinfo{pages}{2808--2815} (\bibinfo{year}{2023}).

\bibitem{chen2024effect}
\bibinfo{author}{Chen, S.} \emph{et~al.}
\newblock \bibinfo{journal}{\bibinfo{title}{The effect of using a large language model to respond to patient messages}}.
\newblock {\emph{\JournalTitle{The Lancet Digital Health}}} \textbf{\bibinfo{volume}{6}}, \bibinfo{pages}{e379--e381} (\bibinfo{year}{2024}).

\bibitem{gandhi2023can}
\bibinfo{author}{Gandhi, T.~K.} \emph{et~al.}
\newblock \bibinfo{journal}{\bibinfo{title}{How can artificial intelligence decrease cognitive and work burden for front line practitioners?}}
\newblock {\emph{\JournalTitle{JAMIA open}}} \textbf{\bibinfo{volume}{6}}, \bibinfo{pages}{ooad079} (\bibinfo{year}{2023}).

\bibitem{baxter2024generative}
\bibinfo{author}{Baxter, S.~L.}, \bibinfo{author}{Longhurst, C.~A.}, \bibinfo{author}{Millen, M.}, \bibinfo{author}{Sitapati, A.~M.} \& \bibinfo{author}{Tai-Seale, M.}
\newblock \bibinfo{journal}{\bibinfo{title}{Generative artificial intelligence responses to patient messages in the electronic health record: early lessons learned}}.
\newblock {\emph{\JournalTitle{JAMIA open}}} \textbf{\bibinfo{volume}{7}}, \bibinfo{pages}{ooae028} (\bibinfo{year}{2024}).

\bibitem{small2024large}
\bibinfo{author}{Small, W.~R.} \emph{et~al.}
\newblock \bibinfo{journal}{\bibinfo{title}{Large language model--based responses to patients’ in-basket messages}}.
\newblock {\emph{\JournalTitle{JAMA network open}}} \textbf{\bibinfo{volume}{7}}, \bibinfo{pages}{e2422399--e2422399} (\bibinfo{year}{2024}).

\bibitem{eriksen2023use}
\bibinfo{author}{Eriksen, A.~V.}, \bibinfo{author}{M{\"o}ller, S.} \& \bibinfo{author}{Ryg, J.}
\newblock \bibinfo{title}{Use of gpt-4 to diagnose complex clinical cases} (\bibinfo{year}{2023}).

\bibitem{nori2023capabilities}
\bibinfo{author}{Nori, H.}, \bibinfo{author}{King, N.}, \bibinfo{author}{McKinney, S.~M.}, \bibinfo{author}{Carignan, D.} \& \bibinfo{author}{Horvitz, E.}
\newblock \bibinfo{journal}{\bibinfo{title}{Capabilities of gpt-4 on medical challenge problems}}.
\newblock {\emph{\JournalTitle{arXiv preprint arXiv:2303.13375}}}  (\bibinfo{year}{2023}).

\bibitem{hao2024advancing}
\bibinfo{author}{Hao, Y.}, \bibinfo{author}{Liu, Z.}, \bibinfo{author}{Riter, R.~N.} \& \bibinfo{author}{Kalantari, S.}
\newblock \bibinfo{title}{Advancing patient-centered shared decision-making with ai systems for older adult cancer patients}.
\newblock In \emph{\bibinfo{booktitle}{Proceedings of the CHI Conference on Human Factors in Computing Systems}}, \bibinfo{pages}{1--20} (\bibinfo{year}{2024}).

\bibitem{holmes2023evaluating}
\bibinfo{author}{Holmes, J.} \emph{et~al.}
\newblock \bibinfo{journal}{\bibinfo{title}{Evaluating large language models on a highly-specialized topic, radiation oncology physics}}.
\newblock {\emph{\JournalTitle{Frontiers in Oncology}}} \textbf{\bibinfo{volume}{13}}, \bibinfo{pages}{1219326} (\bibinfo{year}{2023}).

\bibitem{garcia2024artificial}
\bibinfo{author}{Garcia, P.} \emph{et~al.}
\newblock \bibinfo{journal}{\bibinfo{title}{Artificial intelligence--generated draft replies to patient inbox messages}}.
\newblock {\emph{\JournalTitle{JAMA Network Open}}} \textbf{\bibinfo{volume}{7}}, \bibinfo{pages}{e243201--e243201} (\bibinfo{year}{2024}).

\bibitem{rezayi2022}
\bibinfo{author}{Rezayi, S.} \emph{et~al.}
\newblock \bibinfo{title}{Clinicalradiobert: Knowledge-infused few shot learning for clinical notes named entity recognition}.
\newblock In \emph{\bibinfo{booktitle}{Machine Learning in Medical Imaging: 13th International Workshop, MLMI 2022, Held in Conjunction with MICCAI 2022, Singapore, September 18, 2022, Proceedings}}, \bibinfo{pages}{269–278}, \doiprefix\url{10.1007/978-3-031-21014-3_28} (\bibinfo{publisher}{Springer-Verlag}, \bibinfo{address}{Berlin, Heidelberg}, \bibinfo{year}{2022}).

\bibitem{wu2024}
\bibinfo{author}{Wu, Z.} \emph{et~al.}
\newblock \bibinfo{journal}{\bibinfo{title}{Exploring the trade-offs: Unified large language models vs local fine-tuned models for highly-specific radiology nli task}}.
\newblock {\emph{\JournalTitle{IEEE Transactions on Big Data}}} \doiprefix\url{10.48550/arXiv.2304.09138} (\bibinfo{year}{2024}).
\newblock \bibinfo{note}{Accepted}.

\bibitem{LIAO2024128576}
\bibinfo{author}{Liao, W.} \emph{et~al.}
\newblock \bibinfo{journal}{\bibinfo{title}{Mask-guided bert for few-shot text classification}}.
\newblock {\emph{\JournalTitle{Neurocomputing}}} \textbf{\bibinfo{volume}{610}}, \bibinfo{pages}{128576}, \doiprefix\url{https://doi.org/10.1016/j.neucom.2024.128576} (\bibinfo{year}{2024}).

\bibitem{HOLMES2024}
\bibinfo{author}{Holmes, J.} \emph{et~al.}
\newblock \bibinfo{journal}{\bibinfo{title}{Benchmarking a foundation large language model on its ability to relabel structure names in accordance with the american association of physicists in medicine task group-263 report}}.
\newblock {\emph{\JournalTitle{Practical Radiation Oncology}}} \doiprefix\url{https://doi.org/10.1016/j.prro.2024.04.017} (\bibinfo{year}{2024}).

\bibitem{liu2023tailoring}
\bibinfo{author}{Liu, Z.} \emph{et~al.}
\newblock \bibinfo{title}{Tailoring large language models to radiology: A preliminary approach to llm adaptation for a highly specialized domain}.
\newblock In \emph{\bibinfo{booktitle}{Machine Learning in Medical Imaging: 14th International Workshop, MLMI 2023, Held in Conjunction with MICCAI 2023, Vancouver, BC, Canada, October 8, 2023, Proceedings, Part I}}, \bibinfo{pages}{464–473}, \doiprefix\url{10.1007/978-3-031-45673-2_46} (\bibinfo{publisher}{Springer-Verlag}, \bibinfo{address}{Berlin, Heidelberg}, \bibinfo{year}{2023}).

\bibitem{LIU2023100045}
\bibinfo{author}{Liu, C.} \emph{et~al.}
\newblock \bibinfo{journal}{\bibinfo{title}{Artificial general intelligence for radiation oncology}}.
\newblock {\emph{\JournalTitle{Meta-Radiology}}} \textbf{\bibinfo{volume}{1}}, \bibinfo{pages}{100045}, \doiprefix\url{https://doi.org/10.1016/j.metrad.2023.100045} (\bibinfo{year}{2023}).

\bibitem{Dai2023ChatAugLC}
\bibinfo{author}{Dai, H.} \emph{et~al.}
\newblock \bibinfo{journal}{\bibinfo{title}{Chataug: Leveraging chatgpt for text data augmentation}}.
\newblock {\emph{\JournalTitle{IEEE Transsations on Big Data}}}  (\bibinfo{year}{2024}).
\newblock \bibinfo{note}{Accepted}.

\bibitem{XIAO2024102204}
\bibinfo{author}{Xiao, Z.} \emph{et~al.}
\newblock \bibinfo{journal}{\bibinfo{title}{Instruction-vit: Multi-modal prompts for instruction learning in vision transformer}}.
\newblock {\emph{\JournalTitle{Information Fusion}}} \textbf{\bibinfo{volume}{104}}, \bibinfo{pages}{102204}, \doiprefix\url{https://doi.org/10.1016/j.inffus.2023.102204} (\bibinfo{year}{2024}).

\bibitem{liu2023radoncgptlargelanguagemodel}
\bibinfo{author}{Liu, Z.} \emph{et~al.}
\newblock \bibinfo{title}{Radonc-gpt: A large language model for radiation oncology} (\bibinfo{year}{2023}).
\newblock \eprint{2309.10160}.

\bibitem{chang2024survey}
\bibinfo{author}{Chang, Y.} \emph{et~al.}
\newblock \bibinfo{journal}{\bibinfo{title}{A survey on evaluation of large language models}}.
\newblock {\emph{\JournalTitle{ACM Transactions on Intelligent Systems and Technology}}} \textbf{\bibinfo{volume}{15}}, \bibinfo{pages}{1--45} (\bibinfo{year}{2024}).

\bibitem{iroju2015systematic}
\bibinfo{author}{Iroju, O.~G.} \& \bibinfo{author}{Olaleke, J.~O.}
\newblock \bibinfo{journal}{\bibinfo{title}{A systematic review of natural language processing in healthcare}}.
\newblock {\emph{\JournalTitle{International Journal of Information Technology and Computer Science}}} \textbf{\bibinfo{volume}{8}}, \bibinfo{pages}{44--50} (\bibinfo{year}{2015}).

\bibitem{liu2024towards}
\bibinfo{author}{Liu, L.} \emph{et~al.}
\newblock \bibinfo{title}{Towards automatic evaluation for llms' clinical capabilities: Metric, data, and algorithm}.
\newblock In \emph{\bibinfo{booktitle}{Proceedings of the 30th ACM SIGKDD Conference on Knowledge Discovery and Data Mining}}, \bibinfo{pages}{5466--5475} (\bibinfo{year}{2024}).

\bibitem{abeysinghe2024challenges}
\bibinfo{author}{Abeysinghe, B.} \& \bibinfo{author}{Circi, R.}
\newblock \bibinfo{journal}{\bibinfo{title}{The challenges of evaluating llm applications: An analysis of automated, human, and llm-based approaches}}.
\newblock {\emph{\JournalTitle{arXiv preprint arXiv:2406.03339}}}  (\bibinfo{year}{2024}).

\bibitem{wei2024evaluation}
\bibinfo{author}{Wei, Q.} \emph{et~al.}
\newblock \bibinfo{journal}{\bibinfo{title}{Evaluation of chatgpt-generated medical responses: a systematic review and meta-analysis}}.
\newblock {\emph{\JournalTitle{Journal of Biomedical Informatics}}} \bibinfo{pages}{104620} (\bibinfo{year}{2024}).

\bibitem{maccartney2009natural}
\bibinfo{author}{MacCartney, B.}
\newblock \emph{\bibinfo{title}{Natural language inference}} (\bibinfo{publisher}{Stanford University}, \bibinfo{year}{2009}).

\bibitem{miller1991contextual}
\bibinfo{author}{Miller, G.~A.} \& \bibinfo{author}{Charles, W.~G.}
\newblock \bibinfo{journal}{\bibinfo{title}{Contextual correlates of semantic similarity}}.
\newblock {\emph{\JournalTitle{Language and cognitive processes}}} \textbf{\bibinfo{volume}{6}}, \bibinfo{pages}{1--28} (\bibinfo{year}{1991}).

\bibitem{flesch1948new}
\bibinfo{author}{Flesch, R.}
\newblock \bibinfo{journal}{\bibinfo{title}{A new readability yardstick.}}
\newblock {\emph{\JournalTitle{Journal of applied psychology}}} \textbf{\bibinfo{volume}{32}}, \bibinfo{pages}{221} (\bibinfo{year}{1948}).

\bibitem{jindal2017assessing}
\bibinfo{author}{Jindal, P.} \& \bibinfo{author}{MacDermid, J.~C.}
\newblock \bibinfo{journal}{\bibinfo{title}{Assessing reading levels of health information: uses and limitations of flesch formula}}.
\newblock {\emph{\JournalTitle{Education for Health}}} \textbf{\bibinfo{volume}{30}}, \bibinfo{pages}{84--88} (\bibinfo{year}{2017}).

\bibitem{cognetta_rieke_2024}
\bibinfo{author}{Cognetta-Rieke, C.}
\newblock \bibinfo{journal}{\bibinfo{title}{Mayo clinic department of nursing leveraging artificial intelligence for automating draft patient message responses}}.
\newblock {\emph{\JournalTitle{Mayo AI Summit 2024}}} .

\end{thebibliography}

\newpage
\section{Appendix}
\subsection{Grading Study Details}

\begin{figure}[ht]
  \centering
  \includegraphics[width=19cm]{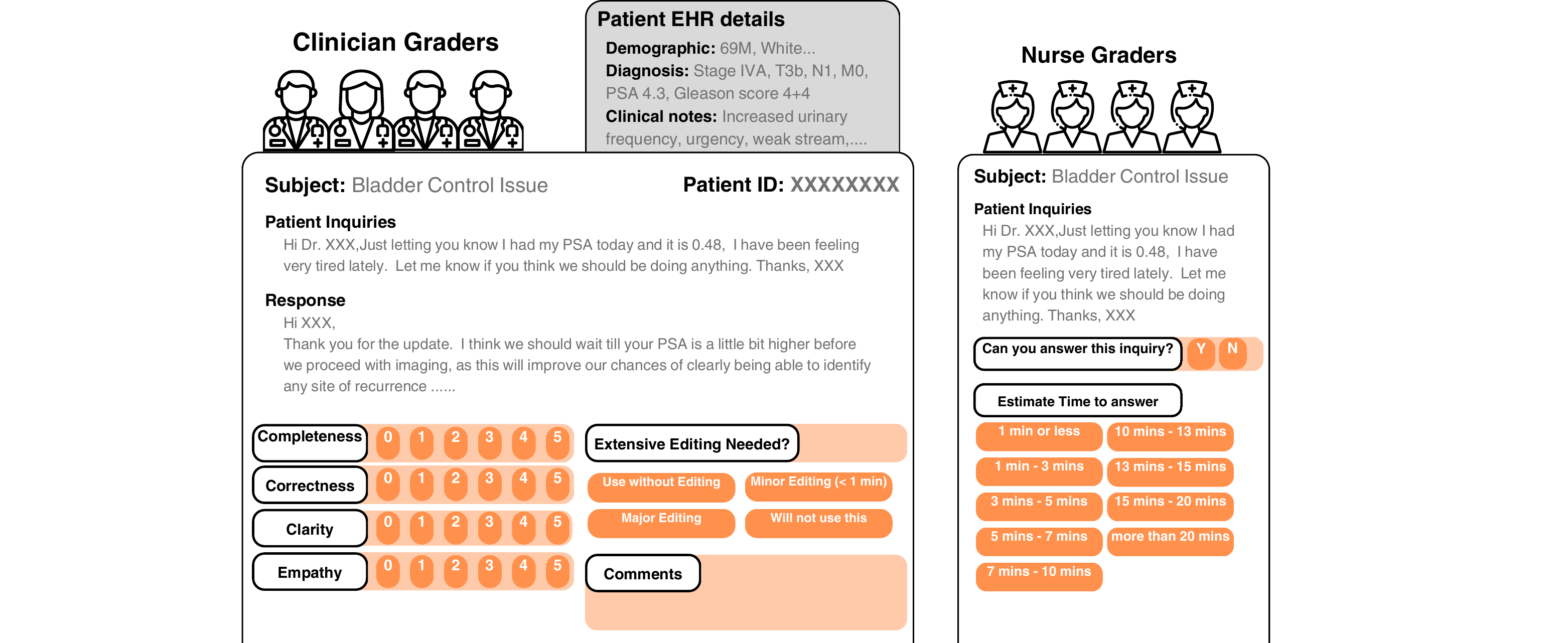}
  \caption{In-Basket Message Grading Study Details and Sample GUI.} 
  \label{Grading Study Details}
\end{figure}

\subsection{Prompt Engineering}
\label{prompt}
From the instructions above, we tested several different prompts and finally used this as our final prompt: "\textit{Patient \underline{\#ID} has sent an in-basket message. Please generate a response to their message. Before generating the response, first retrieve the patient details, patient treatment details, patient diagnosis details, and patient clinical notes. Do not pull the in-basket messages. Retrieve all types of patient data simultaneously. In writing your response, feel free to make recommendations as if you were the attending physician (since your response will be approved by the attending physician). When handling prescriptions, prioritize over-the-counter if appropriate. Sign off as the attending physician. Do not mention that the patient should contact their provider, since you are acting as their provider. Prior to giving your response, explain your reasoning step by step in an analysis section. As part of your analysis, indicate whether the patient has provided enough information to adequately respond to the message. If you determine that the patient has not provided enough information, please ask for more information in your message to the patient. Assume the patient has a high school education level. Here is the in-basket message that you should respond to: \underline{Message Details}.}"

\newpage
\subsection{In-Basket Messages Grading Rubric}
\label{rubric}
\textbf{0 - Not applicable; 1 - Strongly disagree; 2 - Disagree; 3 - Neutral; 4 - Agree; 5 - Strongly agree}

\subsection*{Completeness (6-point scale)}
\vspace{1mm}
\textbf{Definition:} The extent to which the response addresses all parts of the patient's message.

\vspace{0.2mm}
\textbf{Key Points:}
\begin{itemize}
    \item Does the response cover all the questions or concerns raised by the patient?
    \item Are there any missing elements that the patient might need to know?
    \item Does the response provide a thorough and comprehensive answer?
\end{itemize}

\subsection*{Correctness}
\vspace{1mm}
\textbf{Definition:} The accuracy and reliability of the information provided in the response.

\vspace{0.2mm}
\textbf{Key Points:}
\begin{itemize}
    \item Is the information factually correct?
    \item Are medical terms and treatment options accurately described?
    \item Does the response avoid any misleading or incorrect statements?
\end{itemize}

\subsection*{Clarity}
\vspace{1mm}
\textbf{Definition:} The ease with which the patient can understand the response.

\vspace{0.2mm}
\textbf{Key Points:}
\begin{itemize}
    \item How clear and easy is the language to understand?
    \item Are medical terms explained in a way that a layperson can comprehend?
    \item Is the response well-organized and logically structured?
\end{itemize}

\subsection*{Empathy}
\vspace{1mm}
\textbf{Definition:} The degree to which the response shows understanding and sensitivity to the patient's emotional state.

\vspace{0.2mm}
\textbf{Key Points:}
\begin{itemize}
    \item Does the response acknowledge the patient's feelings and concerns?
    \item Is the tone compassionate and supportive?
    \item Does the response make the patient feel heard and cared for?
\end{itemize}

\subsection*{Extensive Editing (4-point scale)}
\vspace{1mm}
\textbf{Definition:} The degree to which the response can be sent to the patient directly.

\vspace{0.2mm}
\textbf{Scale:}
\begin{itemize}
    \item Would use this without editing.
    \item Minor (<1 minute) editing.
    \item Major editing.
    \item Would not use this.
\end{itemize}

\newpage

\subsection{Grader Study Result Statistics}
\label{statistics}
\begin{table*}[h!]
    \centering
    \renewcommand{\arraystretch}{1.1}{ 
    \centering
    \begin{tabular}{c|c|c|c|c}
    \toprule[2pt]
 \textbf {Metrics} & \textbf{Completeness} & \textbf{Correctness} & \textbf{Clarity} & \textbf{Empathy}\\
    \midrule[2pt]
    \centering
    Two Graders Mean (SD) of RadOnc-GPT Responses & 4.56 (0.85) & 4.33 (1.05) &  4.91 (0.43) & 4.71 (0.57)\\     
    Two Graders Mean (SD) of Human Care Team Responses & 4.65 (0.66) & 4.69 (0.67) &  4.90 (0.43) & 4.52 (0.75)\\
    t-statistics & -3.96 & -1.08 & 0 & -4.23\\
    \textit{p}-Value & < 0.01 & > 0.05 & > 0.05 & < 0.01\\
    Wilcoxon signed-rank test & 1033.0 & 2168.0 & 196.0 & 854.0 \\
    R-Squared & 0.30 & 0.40 & 0.23 & 0.31\\
    ICC & 0.68 & 0.77 & 0.65 & 0.67\\
    Pearson Correlation Coefficient & 0.55 & 0.63 & 0.48 & 0.56\\
    Cohen's Kappa Score & 0.27 & 0.30 & 0.21 & 0.32\\
    Three Graders' Mean Variability [41 messages] & 0.94 & 0.98 & 0.74 & 0.74\\
    \bottomrule[2pt]
    \end{tabular}
    } 
    \vspace{0.2cm}
    \caption{Clinician Grader Result Statistics.}
    \label{tab:clinician_table}
\end{table*}

\subsection{More Results}

\subsubsection{Grader Bias}
The correlation of 0.717 indicates a strong positive relationship between the results of the two graders. This means that when one grader gives a higher score, the other grader tends to give a higher score as well, showing that their evaluations are largely consistent and in agreement. Small T-test and >0.05 \textit{P}-Value indicate that there is no statistically significant difference between the results of the two graders, meaning their assessments are generally consistent with each other.
To understand the bias between two independent graders, we used a Radar Chart (Figure \ref{Bias} Right) to understand the differences between the four categories. 

\begin{figure}[h!]
  \centering
  \includegraphics[width=18cm]{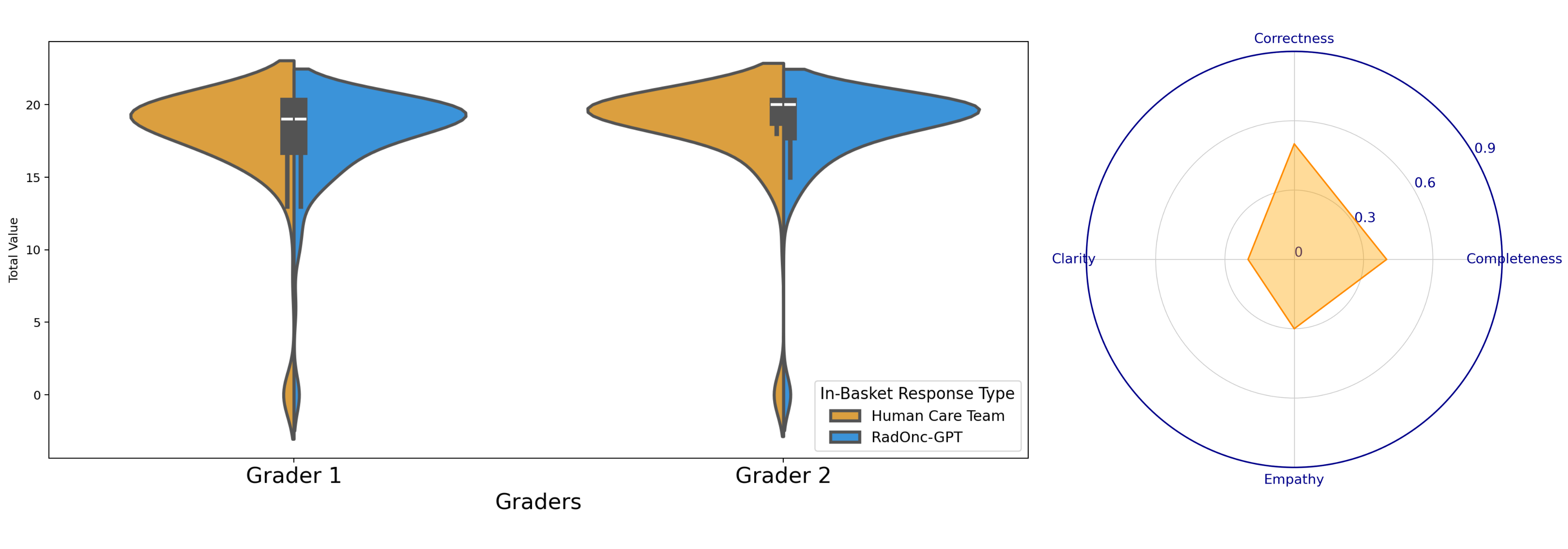}
  \caption{Comparisons Between Two Graders' Scorings on Four Categories: \textit{Completeness, Correctness, Clarity, and Empathy}.}
  \label{Bias}
\end{figure}

The third grader graded 41 responses when the differences between two graders are greater than one. The fourth grader graded 2 responses when third grader still introduced big differences.


The analysis of "\textit{Completeness}," "\textit{Correctness}," "\textit{Clarity}," and "\textit{Empathy}" scores between two clinician graders revealed notable differences in certain areas. In the category of "\textit{Completeness}," the t-test yielded a t-statistic of -3.960 with \textit{p} < 0.05, indicating a statistically significant difference between the two graders. Similarly, in "\textit{Empathy}," a t-statistic of -4.235 with \textit{p} < 0.05, highlighting a significant difference in graders' perspectives interacting with patients. On the other hand, the "\textit{Correctness}" category presents a different picture. The t-test of -1.076 with \textit{p} > 0.05 suggest no statistically significant difference between the two clinicians, indicating that both clinician graders are similarly accurate in their assessments. Moreover, the "\textit{Clarity}" scores show complete parity between the two clinicians, with a t-statistic of 0.0 and \textit{p}-value > 0.05, meaning their clarity in communication is identical.


\begin{figure}[h]
  \centering
  \includegraphics[width=11cm]{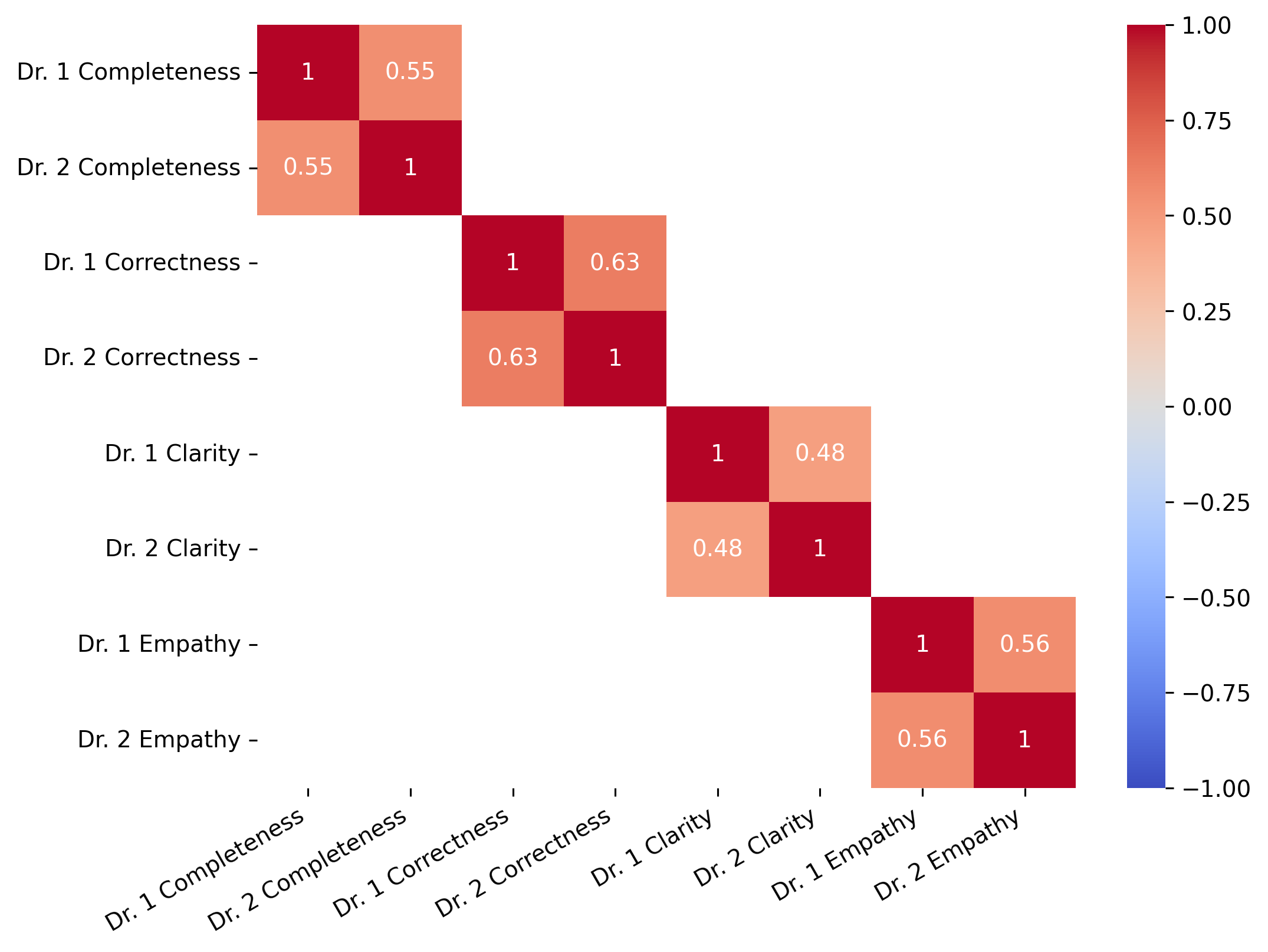}
  \caption{Heatmap of Two Clinician Graders' Correlation Across Four Categories.}
  \label{Bias}
\end{figure}

\subsection{Analysis of Qualitative Comments from Clinician Graders}

\begin{figure}[h]
  \centering
  \includegraphics[width=13.5cm]{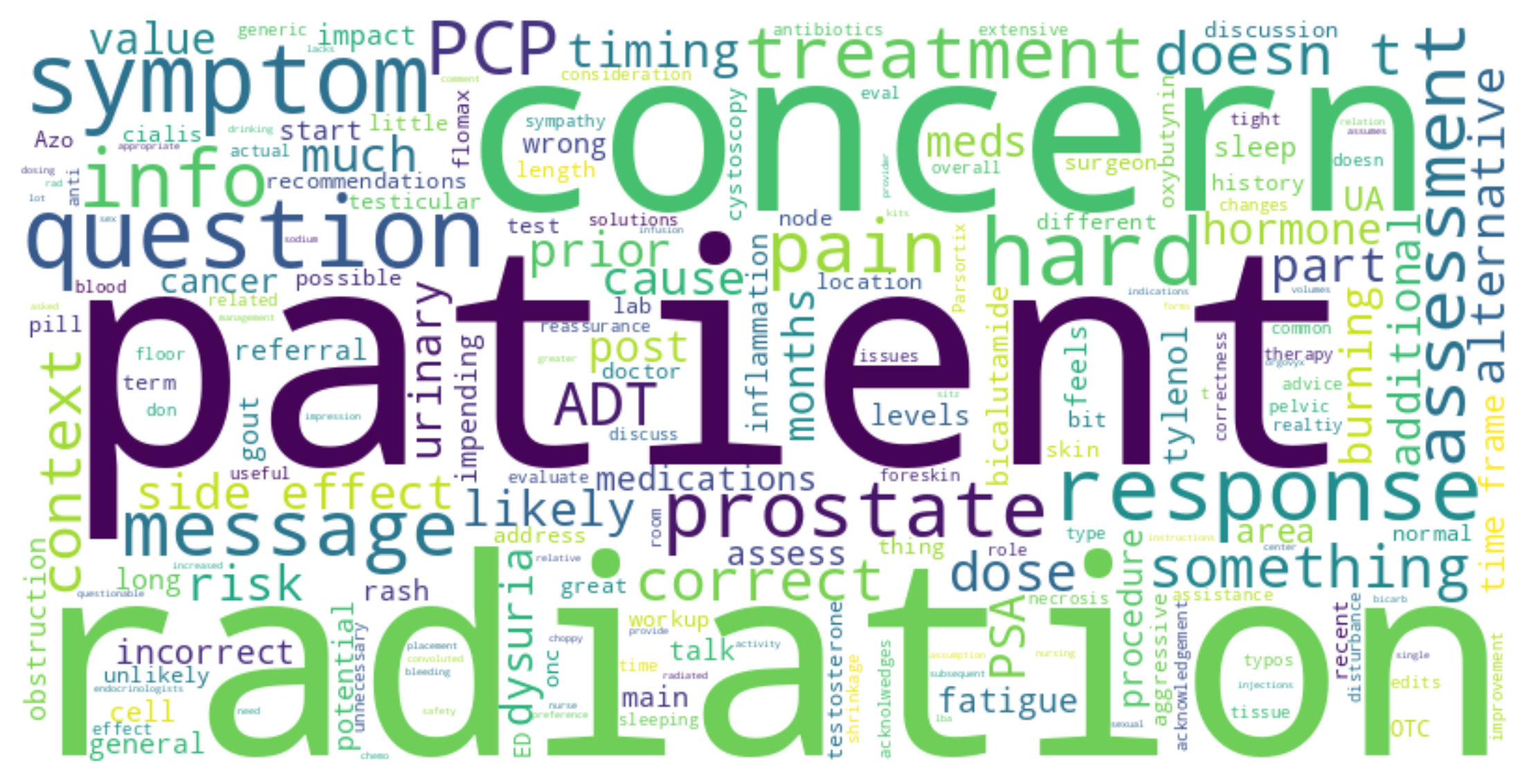}
  \caption{Wordcloud of Qualitative Comments from Clinician Graders}
  \label{Wordcloud}
\end{figure}

Clinician graders optionally provided comments, which indicated a focus on patient concerns, with words like "patient," "radiation," "prostate," and "symptoms" standing out (Figure \ref{Wordcloud}). Issues related to treatment side effects, proper assessment, and clarity in responses were frequently mentioned, highlighting the importance of addressing specific patient needs and improving communication.

\subsection{LLM Graders}
We also conducted a study to evaluate whether different LLMs could match the evaluation quality of clinician graders. Using the same study design as for the clinicians—randomized single-blinded grading, the same grading rubrics, and independent grading without knowing the clinicians' scores—we tested GPT-4o and Gemini on four criteria: "\textit{Completeness}," "\textit{Correctness}," "\textit{Clarity}," and "\textit{Empathy}". The LLMs were integrated with the EHR and provided with the same grading rubrics as prompts used by the clinician graders (Appendix \ref{rubric}).

\begin{figure}[h!]
  \centering
  \includegraphics[width=12.3cm]{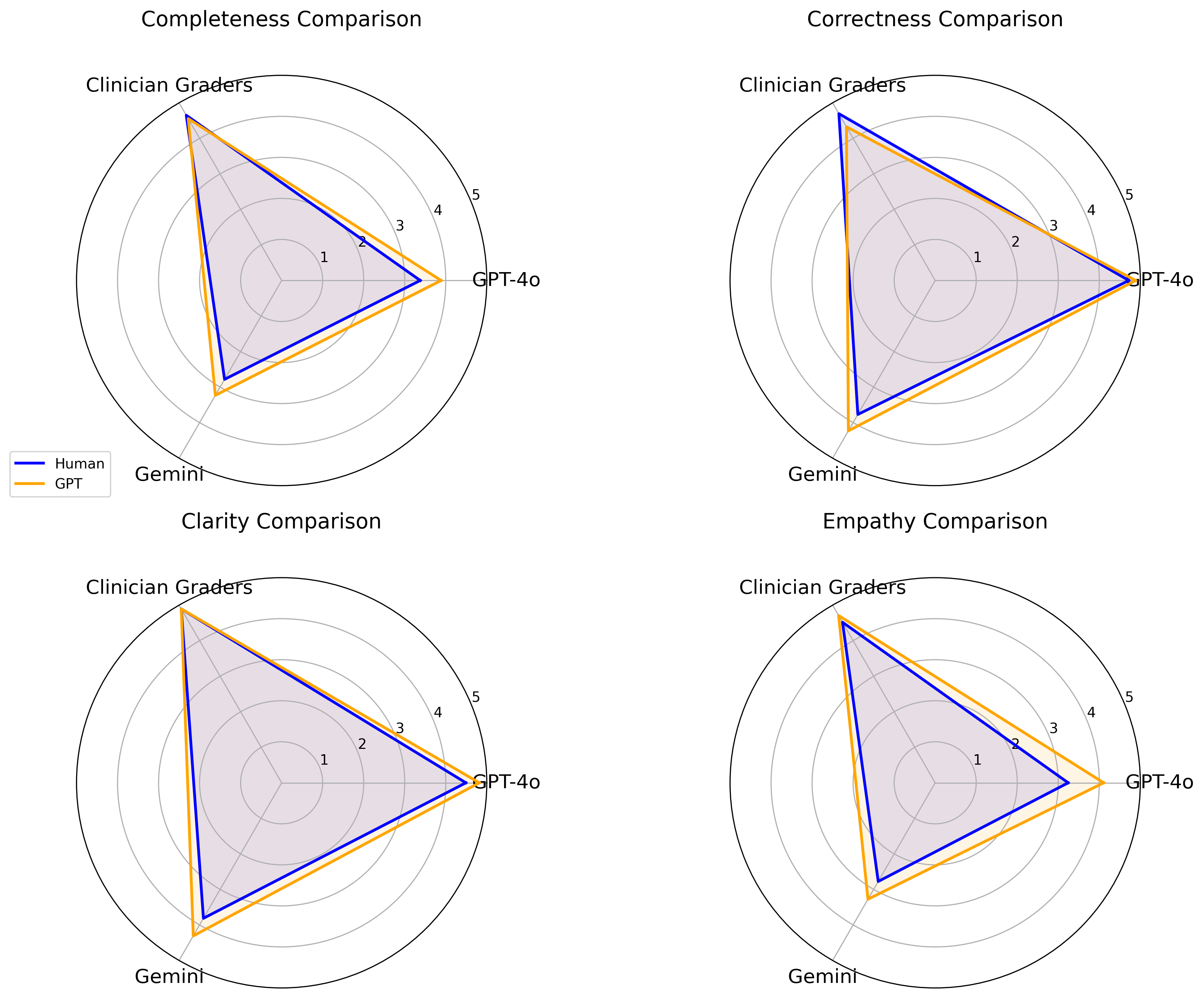}
  \caption{Radar Charts to compare four dimensions with three different graders.}
  \label{graders}
\end{figure}

Under a zero-shot prompt engineering approach, GPT-4o rated RadOnc-GPT responses at 17.70 and clinical care team responses at 15.83, while Gemini rated RadOnc-GPT responses at 15.05 and clinical care team responses at 13.14, all out of 20. The LLM graders generally favored RadOnc-GPT responses but rated all responses lower overall compared to clinician graders.  This discrepancy may be due to factors such as the format preferred by LLMs, an unclear grading rubric, or the need for more domain-specific or clinical knowledge. Although the study was randomized and single-blinded, the LLMs likely recognized whether the responses were generated by RadOnc-GPT or the human care team, showing a preference for RadOnc-GPT responses. The detailed results for each category are displayed in Table \ref{LLM_Grader_Stats} and the radar chart Figure \ref{graders}.

\begin{table}[htbp]
\centering
\begin{tabular}{l|l|c|c|c}
\toprule[2pt]
\textbf{Category} & \textbf{Response Type} & \textbf{Clinician Graders} & \textbf{GPT-4o} & \textbf{Gemini} \\ 
    \midrule[2pt]
\textbf{Completeness} & RadOnc-GPT   & 4.55 (0.77) & 3.89 (0.84) & 3.23 (0.89) \\ \cline{2-5} 
                      & Clinical Care Team & 4.65 (0.59) & 3.38 (0.91) & 2.78 (0.88) \\ \hline
\textbf{Correctness}  & RadOnc-GPT   & 4.32 (0.94) & 4.89 (0.40) & 4.23 (0.70) \\ \cline{2-5} 
                      & Clinical Care Team  & 4.69 (0.58) & 4.72 (0.58) & 3.77 (0.87) \\ \hline
\textbf{Clarity}      & RadOnc-GPT   & 4.89 (0.45) & 4.82 (0.43) & 4.31 (0.64) \\ \cline{2-5} 
                      & Clinical Care Team  & 4.89 (0.41) & 4.49 (0.73) & 3.81 (0.88) \\ \hline
\textbf{Empathy}      & RadOnc-GPT   & 4.71 (0.49) & 4.11 (0.98) & 3.28 (0.90) \\ \cline{2-5} 
                      & Clinical Care Team  & 4.52 (0.67) & 3.24 (1.19) & 2.78 (0.98) \\ 
    \bottomrule[2pt]
\end{tabular}
\caption{Mean and Standard Deviation of Scores for Clinician Graders, GPT-4o, and Gemini}
\label{LLM_Grader_Stats}
\end{table}


Research into the evaluation process with GPT-4o and Gemini showed that it is feasible to train LLMs to serve as evaluators on par with human experts. As LLMs become more aligned with clinician graders, they could also improve the quality of their own responses. However, existed LLMs relying solely on zero-shot prompts, without proper domain-specific training and guidance, still require further refinement.

\end{document}